\documentclass[letterpaper, 10 pt, conference]{template/ieeeconf}  % Comment this line out if you need a4paper

\IEEEoverridecommandlockouts                              % This command is only needed if 
\usepackage{amsmath} % assumes amsmath package installed
\usepackage{amssymb}  % assumes amsmath package installed
\usepackage{graphicx}
\usepackage[caption=false]{subfig}
\usepackage{amsmath} % assumes amsmath package installed
\usepackage{amssymb}  % assumes amsmath package installed
\usepackage{hyphenat}
\usepackage[table]{xcolor}
\usepackage[font=small,belowskip=0pt, aboveskip=6pt]{caption}
\usepackage{multirow}
\usepackage{url}
\usepackage{booktabs,amsfonts,dcolumn}
\usepackage{tabularx}
\usepackage{tabu}
\usepackage{gensymb}
\usepackage{makecell}
\usepackage{hyperref}
\usepackage{comment}

\usepackage{pgfplots}
\usetikzlibrary{pgfplots.statistics, pgfplots.colorbrewer, pgfplots.groupplots, shapes.geometric, decorations.markings, arrows, decorations.text}
\usepgfplotslibrary{fillbetween}
\usepackage{pgfplotstable}
\pgfplotsset{compat=1.15}
\usetikzlibrary{external}
\usetikzlibrary{positioning, calc, fit}
\usetikzlibrary{arrows.meta,arrows}
\usepackage{balance}

\usepackage[style=ieee,mincitenames=1,maxcitenames=2,maxbibnames=10,doi=false,isbn=false,url=false,eprint=false]{biblatex}
\DeclareSourcemap{
  \maps[datatype=bibtex, overwrite]{
    \map{
      \step[fieldset=address, null]
      \step[fieldset=location, null]
    }
  }
}

\usepackage{xspace}

\long\def\invis#1{}

\newcommand\eq[1]{Eq.~\eqref{#1}}
\newcommand\fig[1]{Fig.~\ref{#1}}

\makeatletter
\DeclareRobustCommand\onedot{\futurelet\@let@token\@onedot}
\def\@onedot{\ifx\@let@token.\else.\null\fi\xspace}

\makeatother

\usepackage[group-separator={,}]{siunitx}

\title{\LARGE \bf
Stereo-NEC: Enhancing Stereo Visual-Inertial SLAM Initialization with Normal Epipolar Constraints}

\author{Weihan Wang$^{a,}$$^b$, Chieh Chou$^b$, 
Ganesh Sevagamoorthy$^b$, Kevin Chen$^b$, Zheng Chen$^b$,\\ Ziyue Feng$^b$, Youjie Xia$^b$, Feiyang Cai$^c$, 
Yi Xu$^b$, Philippos Mordohai$^a$
\thanks{$^a$Stevens Institute of Technology, Hoboken, NJ, USA, 07030, {\tt\small \{wwang103,pmordoha\}@stevens.edu}}
\thanks{$^b$OPPO US Research Center, Palo Alto, CA, USA, 94303, \tt\small \{weihan.wang, 
		chieh.chou, ganesh.sevagamoorthy, kevin.chen, zheng.chen, ziyue.feng, youjie.xia, yi.xu\}@oppo.com}
\thanks{$^c$Stony Brook University, Stony Brook, NY 11794, {\tt\small feiyang.cai@stonybrook.edu}}
\thanks{
This research has been supported in part by the National Science Foundation under award 2024653.}
 }

\begin{document}

\begin{minipage}{0.90\textwidth}\ \\[12pt]
\vspace{3in}
\begin{center}
     This paper has been accepted for publication in \textit{IEEE Conference on Robotics and Automation 2024}.  
\end{center}
  \vspace{1in}
  ©2024 IEEE. Personal use of this material is permitted. Permission from IEEE must be obtained for all other uses, in any current or future media, including reprinting/republishing this material for advertising or promotional purposes, creating new collective works, for resale or redistribution to servers or lists, or reuse of any copyrighted component of this work in other works.
\end{minipage}

\newpage

\maketitle
\thispagestyle{empty}
\pagestyle{empty}

%%%%%%%%%%%%%%%%%%%%%%%%%%%%%%%%%%%%%%%%%%%%%%%%%%%%%%%%%%%%%%%%%%%%%%%%%%%%%%%%
\begin{abstract}
We propose an accurate and robust initialization approach for stereo visual-inertial SLAM systems. Unlike the current state-of-the-art method, which heavily relies on the accuracy of a pure visual SLAM system to estimate inertial variables without updating camera poses, potentially compromising accuracy and robustness, our approach offers a different solution. 
We realize the crucial impact of precise gyroscope bias estimation on rotation accuracy. This, in turn, affects trajectory accuracy due to the accumulation of translation errors.
To address this, we first independently estimate the gyroscope bias and use it to formulate a maximum a posteriori problem for further refinement. After this refinement, we proceed to update the rotation estimation by performing IMU integration with gyroscope bias removed from gyroscope measurements. We then leverage robust and accurate rotation estimates to enhance translation estimation via 3-DoF bundle adjustment. Moreover, we introduce a novel approach for determining the success of the initialization by evaluating the residual of the normal epipolar constraint.  Extensive evaluations on the EuRoC dataset illustrate that our method excels in accuracy and robustness. It outperforms ORB-SLAM3, the current leading stereo visual-inertial initialization method, in terms of absolute trajectory error and relative rotation error, while maintaining competitive computational speed. Notably, even with 5 keyframes for initialization, our method consistently surpasses the state-of-the-art approach using 10 keyframes in rotation accuracy. The open source code is available at \url{https://github.com/ApdowJN/Stereo-NEC.git}.
\end{abstract}

%%%%%%%%%%%%%%%%%%%%%%%%%%%%%%%%%%%%%%%%%%%%%%%%%%%%%%%%%%%%%%%%%%%%%%%%%%%%%%%%
\section{INTRODUCTION}
The fusion of cameras and Inertial Measurement Units (IMUs) in Visual-Inertial Simultaneous Localization and Mapping (VI-SLAM) presents a cost-effective, low-power solution for robot perception and AR/VR applications. Cameras offer a rich environment representation, while IMUs measure acceleration and angular velocity, ensuring robustness in fast-motion and texture-less scenes. This synergy makes them ideal complements. Compared to monocular VI-SLAM systems, stereo VI-SLAM systems offer the advantages of a baseline with known scale and the capability to reconstruct 3D geometry even without camera motion.

Initialization in VI-SLAM systems is critical because it impacts their accuracy and robustness. VI-SLAM depends on reliable and precise initial estimates for scale, the gravity direction, initial velocity, acceleration, and gyroscope biases. However,  accomplishing this task is challenging, demanding a swift and accurate recovery of observable parameters from visual and inertial data without prior knowledge.

Compared with the extensive research on monocular systems, there are relatively few VIO solutions designed for stereo systems~\cite{okvis, 7801557, inverse_filter, dso_stereo, 6092505, qin2019general, ORBSLAM3TRO}.This is due to the increased computational demands of processing multiple images and stereo matching.
Similar to monocular VI-SLAM initialization methods~\cite{Kneip, Martinelli, Kaiser, DongSi, jointviinit, viorb, weibovio, VINS,inertialonlyinit,EDI}, 
methods for stereo VI-SLAM initialization are also categorized into two types: joint approaches~\cite{DongSi, openvins} and disjoint approaches~\cite{weibo_stereo, ESVIO, qin2019general, ORBSLAM3TRO}.
Joint approaches handle both visual and inertial parameters together by fusing visual observation and  IMU integration.
However, they tend to overlook the gyroscope bias in the closed-form solution, which results in limited accuracy, while they are computationally expensive. On the other hand, disjoint approaches
first independently solve the Structure-from-Motion (SfM) problem and then derive inertial parameters based on camera poses from a pure visual SLAM system. Therefore, the accuracy of these methods relies heavily on the performance of pure visual SLAM. Previous monocular VI-SLAM approaches have been extended to stereo VI-SLAM using a similar disjoint initialization strategy. For instance, VINS-Fusion~\cite{qin2019general}, an extension of VINS-Mono~\cite{VINS}, follows this approach with a slight difference. VINS-Fusion jointly estimates velocity, gravity vector, and scale through visual-inertial bundle adjustment, rather than treating them separately.
Huang et al.~\cite{weibo_stereo} extended their prior method~\cite{weibovio} to stereo VI-SLAM by introducing an additional scale estimate. Similarly, ORB-SLAM3~\cite{ORBSLAM3TRO} applied the same idea~\cite{inertialonlyinit} to their stereo VI-SLAM system.

The accuracy of pure visual SLAM greatly impacts the performance of disjoint methods. However, even in state-of-the-art stereo VI-SLAM systems like ORB-SLAM3, accurate camera trajectory estimation is assumed in scenarios with adequate baseline between consecutive frames and mild rotation. Nevertheless, in challenging situations such as pure or intense rotation, ORB-SLAM3's initialization may result in reduced accuracy and robustness.

To overcome these limitations and enhance initialization accuracy and robustness in challenging scenarios, we propose Stereo-NEC, which leverages insights from our previous work~\cite{EDI} that takes into account the significant impact of gyroscope bias estimation on rotation accuracy and considers the connection between inertial parameters and visual observations. 
Our approach begins by obtaining accurate rotation estimates, which rely on precise gyroscope bias estimation. This, in turn, plays a crucial role in improving trajectory accuracy by reducing accumulation of translation errors. To achieve accurate gyroscope bias estimation, we extend the concept of monocular normal epipolar constraints (MNEC)~\cite{Frame2Frame, Kneip, Yijia}. MNEC incorporates normal vectors with relative rotation estimation and visual observations, while rotation is approximated using a first-order Taylor series that accounts for the influence of gyroscope bias. Once we have accurate rotation estimates after removing the gyroscope bias, we leverage them to enhance translation estimation by 3-DoF Bundle adjustment.

The main contributions of the proposed initialization method are:
\begin{itemize}
    \item Proposing a new method that utilizes stereo normal epipolar constraints to estimate initial gyroscope bias and uses the latter to initialize a maximum a posteriori (MAP) problem for further refinement.
    \item Enhancing initialization accuracy and robustness by estimating the rotation separately through IMU rotation integration and then utilizing precise and reliable rotation estimates to enhance translation estimation via 3-DoF bundle adjustment.
    \item Introducing a novel approach to assess initialization success by evaluating the residual of the normal epipolar constraint.
\end{itemize}

\section{PRELIMINARIES}
\subsection{Notation}\label{sec:notation}
In this paper, we adopt the following notation: the world frame, body frame, and camera frame are represented by $(\cdot)^\text{w}$, $(\cdot)^\text{b}$, and $(\cdot)^\text{c}$, respectively. For stereo cameras, $\text{c}_\text{L}$ and $\text{c}_\text{R}$ represent the left and right camera, respectively.
We denote rotation matrices with $\textbf{R}$, velocity vectors with $\textbf{v}$,  translation vectors with $\textbf{t}$, and the gravity vector with $\textbf{g} = (0, 0, G)^\top$ and $G$ is the magnitude of gravity.
$\textbf{R}^j_i$ denotes relative rotation from frame $i$ to frame $j$, and  $\textbf{t}^j_i$ denotes relative translation. 
$\text{b}_\text{k}$ is the body frame while taking the $k$-th image, and $\text{c}_\text{k}$ is the left camera frame while taking the $\text{k}$-th image. Acceleration bias and gyroscope bias in the local body frame are represented by $\textbf{b}_\text{a}$ and $\textbf{b}_\text{g}$ respectively. $\boldsymbol\alpha^{\text{b}_\text{k}}_{\text{b}_\text{k+1}}$, $\boldsymbol\beta^{\text{b}_\text{k}}_{\text{b}_\text{k+1}}$, $\boldsymbol\gamma^{\text{b}_\text{k}}_{\text{b}_\text{k+1}}$ represent preintegration of translation, velocity, and rotation from $\text{b}_\text{k}$ to $\text{b}_\text{k+1}$. $\Delta \text{t}_{\text{k}, \text{k+1}}$ denotes the  interval from time $\text{k}$ to time $\text{k+1}$. $\lambda$ represents an eigenvalue of a matrix, and $\lambda_{\min}$ specifically denotes the smallest eigenvalue. $\textbf{n}$ refers to a normal vector. $N$ represents the number of keyframes used for initialization.
\subsection{Monocular Normal Epipolar Constraint}\label{nec}
The monocular normal epipolar constraint (MNEC)~\cite{mnec} encodes the geometric relationship between bearing vectors and the normals of the corresponding epipolar planes defined for two poses of a single mobile camera.
\begin{figure}[!ht]
\begin{center}
\vspace{0.1in}
\includegraphics[width=0.7\columnwidth]{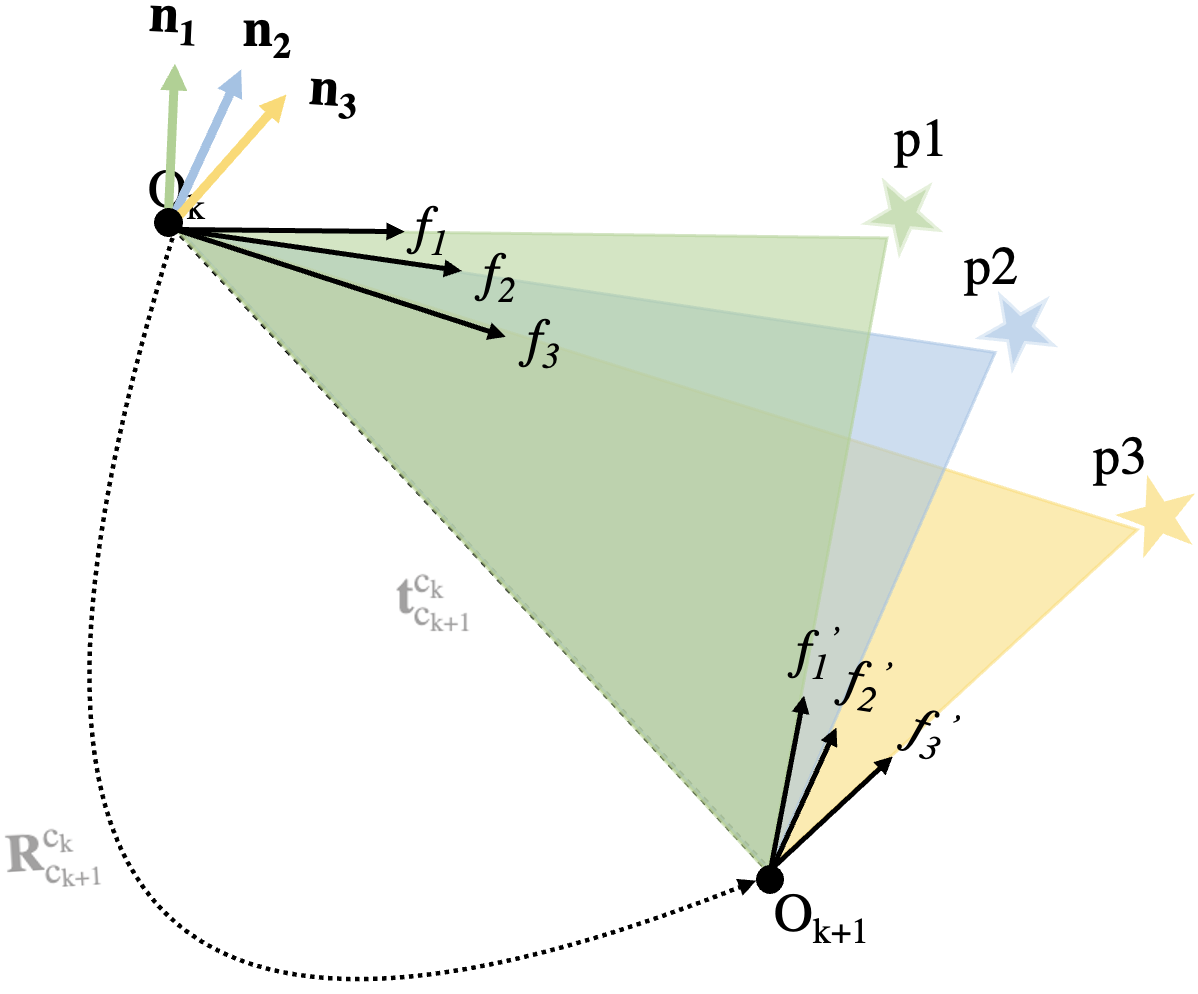}
\end{center}
    \caption{3D points ($\textbf{p}_i$) are represented with different colored five-pointed stars. For each pair of bearing vectors, an epipolar plane is formed (highlighted in green, blue, yellow), and their corresponding normal vectors ($\textbf{n}_i$) are shown in corresponding colors. The normal vectors lie in a same plane. The baseline $\textbf{t}^{
\text{c}_\text{k}}_{\text{c}_\text{k+1}}$ intersects all corresponding epipolar planes.}
    \label{fig:mono_nec}
    % \vspace{-0.07in}x
\end{figure}
As shown in \fig{fig:mono_nec}, when considering feature correspondences in two consecutive times, pairs of unit bearing vectors ($f_i$ and $f_i^{\prime}$) originate from the optical centers ($\text{O}_\text{k}$ and $\text{O}_\text{k+1}$) at time instances \text{k} and \text{k+1}, respectively, and point towards the 3D points ($\textbf{p}_i$). Each pair of bearing vectors define an epipolar plane along with its associated normal vector $\textbf{n}_i = f_i \times \textbf{R}^{\text{c}_\text{k}}_{\text{c}_\text{k+1}}f_i^{\prime}$. The intersection of these epipolar planes forms a line, which is the baseline $\textbf{t}^{
\text{c}_\text{k}}_{\text{c}_\text{k+1}}$ between two frames (marked by the dashed line). The normal vectors collectively form the corresponding epipolar normal plane, which is perpendicular to the baseline $\textbf{t}^{
\text{c}_\text{k}}_{\text{c}_\text{k+1}}$. Assuming that n 
% \xuyi{n have different meanings in II.A. }
% \jay{I think it's ok since one is vector form, one is capital, and one is lower case}
3D points are observed, we stack the n normal vectors of epipolar planes into a matrix $\textbf{N} = \left[\textbf{n}_1...\textbf{n}_n\right]$. The requirement for coplanarity is mathematically equivalent to the minimum eigenvalue of the matrix $\textbf{M} = \textbf{N}\textbf{N}^{\top}$ being zero. 

The residual of the MNEC is given by: 
\begin{equation*}\small
\textit{e}_i= \left|\textbf{n}^{\top}_i\textbf{t}^{
\text{c}_\text{k}}_{\text{c}_\text{k+1}}\right|.
\end{equation*}
There are two applications of MNEC:
\subsubsection{Rotation}
Kneip and Lynen~\cite{Frame2Frame} aim to determine the relative rotation $\textbf{R}^{\text{c}_\text{k}}_{\text{c}_\text{k+1}}$ that minimizes the smallest eigenvalue $\lambda_{\min}({\textbf{M}_{k,k+1}})$:

\begin{equation}\small\label{eq:nec_rotation}
\begin{aligned}
  {\textbf{R}^{\text{c}_\text{k}}_{\text{c}_\text{k+1}}}^* &= \arg\min_{\textbf{R}^{\text{c}_\text{k}}_{\text{c}_\text{k+1}}} \lambda_{\min}({\textbf{M}_{\text{k},\text{k+1}}}), \\
\textbf{M}_{\text{k},\text{k+1}} &= \sum_{i=1}^n(f_i \times \textbf{R}^{\text{c}_\text{k}}_{\text{c}_\text{k+1}}f_i^{\prime}) (f_i \times \textbf{R}^{\text{c}_\text{k}}_{\text{c}_\text{k+1}}f_i^{\prime})^{\top}
\end{aligned}
\end{equation}
The matrix ${\textbf{M}_{k,k+1}}$ possesses specific properties: it is real, symmetric, and positive semi-definite. Furthermore, due to the coplanarity constraint on the normal vectors, its rank is 2. Solving \eq{eq:nec_rotation} is achieved using the Levenberg-Marquardt algorithm.
% This method is capable of accurately estimating $\textbf{R}^{\text{c}_\text{k}}_{\text{c}_\text{k+1}}$, even in scenarios involving pure rotation ($\textbf{t}^{
% \text{c}_\text{k}}_{\text{c}_\text{k+1}} = \textbf{0}$).
\subsubsection{Gyroscope Bias} Inspired by Kneip and Lynen's work, He et al.~\cite{Yijia} employ the MNEC to directly optimize gyroscope bias, incorporating image observations and the camera-IMU extrinsic calibration matrix $[\textbf{R}^{\text{c}}_{\text{b}} | \textbf{t}^{\text{c}}_{\text{b}}]$. This modifies the objective function from \eq{eq:nec_rotation} to \eq{eq:nec_bg}:
\begin{equation}\small\label{eq:nec_bg}
\begin{aligned}
  \textbf{b}^{*}_{\text{g}} = \arg \min_{\textbf{b}_{\text{g}}} \lambda_{\min}&({\textbf{M}_{k,k+1}}), \\
\textbf{n}_i = f_i \times \textbf{R}^{\text{c}}_{\text{b}}&\hat{\boldsymbol\gamma}^{\text{b}_\text{k}}_{\text{b}_\text{k+1}}\textbf{R}^{\text{b}}_{\text{c}}f_i^{\prime} \\
\textbf{M}_{k,k+1} = \sum_{i=1}^n(f_i \times \textbf{R}^{\text{c}}_{\text{b}}\hat{\boldsymbol\gamma}^{\text{b}_\text{k}}_{\text{b}_\text{k+1}}\textbf{R}^{\text{b}}_{\text{c}}f_i^{\prime}) & (f_i \times \textbf{R}^{\text{c}}_{\text{b}}\hat{\boldsymbol\gamma}^{\text{b}_\text{k}}_{\text{b}_\text{k+1}}\textbf{R}^{\text{b}}_{\text{c}}f_i^{\prime})^{\top}\\
% \quad \hat{\boldsymbol\gamma}^{\text{b}_\text{k}}_{\text{b}_\text{k+1}} = \boldsymbol\gamma^{\text{b}_\text{k}}_{\text{b}_\text{k+1}}& \otimes \left [\begin{matrix} 
% 1 \\
% \frac{1}{2}\textbf{J}^{\boldsymbol\gamma}_{\textbf{b}_\text{g}}\textbf{b}_{\text{g}}
% \end{matrix} \right]
\quad \hat{\boldsymbol\gamma}^{\text{b}_\text{k}}_{\text{b}_\text{k+1}} = \boldsymbol\gamma^{\text{b}_\text{k}}_{\text{b}_\text{k+1}}&\text{Exp}(\textbf{J}^{\boldsymbol\gamma}_{\textbf{b}_\text{g}}\textbf{b}_{\text{g}})
\end{aligned}
\end{equation}
where, considering the influence of gyroscope bias over the duration between the $\text{k}$-th keyframe and the $\text{(k+1)}$-th keyframe, 
$\hat{\boldsymbol\gamma}^{\text{b}_\text{k}}_{\text{b}_\text{k+1}}$is estimated using a first-order Taylor approximation of $\boldsymbol\gamma^{\text{b}_\text{k}}_{\text{b}_\text{k+1}}$ and $\textbf{J}^{\boldsymbol\gamma}_{\textbf{b}_\text{g}}$represents how the preintegration changes due to a small difference in gyroscope bias estimation.

\section{Proposed Approach}\label{sec:method}
Our method is motivated by the realization that precise gyroscope bias estimation significantly influences rotation accuracy, ultimately affecting trajectory accuracy due to the accumulation of translation errors. 
Building upon our previous work~\cite{EDI}, which employs an Error-state Kalman Filter (ESKF) to estimate gyroscope bias and corrects rotation estimation, we reduce the reliance on pure visual SLAM for rotation estimation. The underlying idea of our method is rooted in realizing the importance of gyroscope bias.
We start by independently estimating gyroscope bias, which then is employed to formulate a MAP problem for further refinement. Following this refinement, we proceed to update the rotation estimation. With a robust and accurate rotation estimation in place, we leverage it to assist translation estimation.
Our method consists of five steps aimed at deriving precise initial values for keyframes' poses and velocities, gravity direction, and IMU biases:
\begin{itemize}
    \item \textbf{Step 0. Pure Visual SLAM}: Obtain the initial keyframe poses from stereo visual-only SLAM.
    \item \textbf{Step 1. Eigenvalue-based Gyroscope Bias Estimator}: Derive the initial gyroscope bias by formulating an eigenvalue minimization problem using both visual, and gyroscope measurements.
    \item \textbf{Step 2. Gyroscope Bias Refinement, Acceleration Bias, Velocity and Gravity Estimator}: Refine gyroscope bias and  estimate keyframes’ velocities, gravity direction and acceleration bias by solving  an inertial-only MAP estimation problem.
    \item \textbf{Step 3. Rotation-Translation-Decoupled Optimization}: Update the camera rotation estimate by integrating gyroscope measurements with the gyroscope bias removed, and optimize the camera translation using 3-DoF BA.
    \item \textbf{Step 4. Joint Visual-Inertial Bundle Adjustment}: 
    Utilize the solution derived from the previous steps as the initial estimate via visual-inertial MAP to obtain optimal estimates for keyframes' poses, keyframes' velocities, 3D points, gravity direction, and IMU biases.
\end{itemize}
In the following subsections, we provide a more detailed explanation of our initialization steps.
\begin{figure}[!ht]
    \vspace{0.2in}
    \centering
\includegraphics[width=0.8\columnwidth]{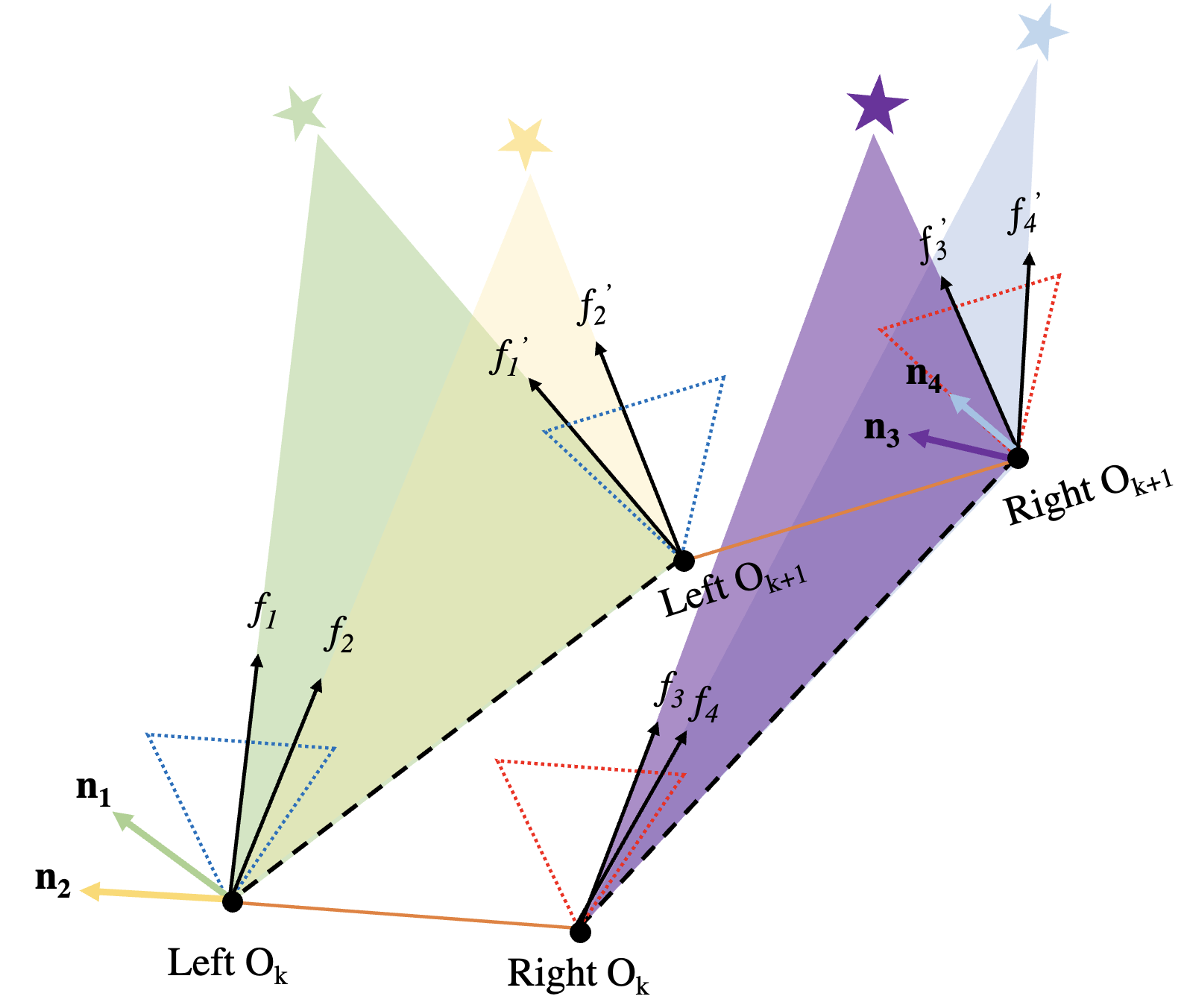}
    \caption{An illustration depicting the geometry of the stereo normal epipolar constraint is shown. 
    The blue and red dashed triangles represent the left and right cameras, respectively. 
    Left $O_\text{k}$ and Left $O_\text{k+1}$ represent the two left optical centers at time $\text{k}$ and $\text{k+1}$ respectively. Similarly, Right $O_\text{k}$ and Right $O_\text{k+1}$ correspond to the two right optical centers. The orange line represents the baseline of the stereo cameras. The temporal epipolar planes of the left camera are colored in green and yellow, while the temporal epipolar planes of the right camera are colored in blue and purple. Each corresponding normal vector ($\textbf{n}_i$) of each temporal epipolar plane is depicted in corresponding colors. Only normal vectors from the same camera are coplanar.
    }
    \label{fig:stereo_nec}
\end{figure}
% For implementation of each step, refer to Section ~\ref{sec:implementation}.
\subsection{Eigenvalue-based Gyroscope Bias Estimator}\label{sec:step1}
As shown in \fig{fig:stereo_nec}, the stereo normal epipolar constraint extends the monocular normal epipolar constraint, thereby enabling the incorporation of additional visual observations to enhance the robustness and accuracy of gyroscope bias estimation. To estimate the initial gyroscope bias while leveraging stereo observation information, we initialize the gyroscope bias by minimizing the smallest eigenvalue $\lambda_{\min}$ in \eq{eq:stereo_nec_bg}:
\begin{equation}\small\label{eq:stereo_nec_bg}
\begin{aligned}
  \textbf{b}^{*}_{\text{g}} = \arg\min_{\textbf{b}_{\text{g}}} \lambda_{\min}, \\
  \lambda_{\min} = \sum_{(\text{k}, \text{k+1})\in \mathcal{E}}(\lambda_{\min}({^\text{L}\textbf{M}_{\text{k},\text{k+1}}}) & + \lambda_{\min}({^\text{R}\textbf{M}_{\text{k},\text{k+1}}}))\\
^\text{L}\textbf{M}_{\text{k},\text{k+1}} = \sum_{i=1}^{n_\text{L}}(f_{{\text{L}}_i} \times \textbf{R}^{\text{c}_{\text{L}}}_{\text{b}}\hat{\boldsymbol\gamma}^{\text{b}_\text{k}}_{\text{b}_\text{k+1}}\textbf{R}^{\text{b}}_{\text{c}_{\text{L}}}f_{{\text{L}}_i}^{\prime}) & (f_{{\text{L}}_i} \times \textbf{R}^{\text{c}_{\text{L}}}_{\text{b}}\hat{\boldsymbol\gamma}^{\text{b}_\text{k}}_{\text{b}_\text{k+1}}\textbf{R}^{\text{b}}_{\text{c}_{\text{L}}}f_{{\text{L}}_i}^{\prime})^{\top}\\
^\text{R}\textbf{M}_{\text{k},\text{k+1}} = \sum_{i=1}^{n_\text{R}}(f_{{\text{R}}_i} \times \textbf{R}^{\text{c}_{\text{R}}}_{\text{b}}\hat{\boldsymbol\gamma}^{\text{b}_\text{k}}_{\text{b}_\text{k+1}}\textbf{R}^{\text{b}}_{\text{c}_{\text{R}}}f_{{\text{R}}_i}^{\prime}) & (f_{{\text{R}}_i} \times \textbf{R}^{\text{c}_{\text{R}}}_{\text{b}}\hat{\boldsymbol\gamma}^{\text{b}_\text{k}}_{\text{b}_\text{k+1}}\textbf{R}^{\text{b}}_{\text{c}_{\text{R}}}f_{{\text{R}}_i}^{\prime})^{\top}\\
\hat{\boldsymbol\gamma}^{\text{b}_\text{k}}_{\text{b}_\text{k+1}} = \boldsymbol\gamma^{\text{b}_\text{k}}_{\text{b}_\text{k+1}}\text{Exp}&(\textbf{J}^{\boldsymbol\gamma}_{\textbf{b}_\text{g}}\textbf{b}_{\text{g}})
\end{aligned}
\end{equation}
where $(f_{{\text{L}}_i}, f_{{\text{L}}_i}^{\prime})$ denotes a pair of bearing vectors in the left camera and  $(f_{{\text{R}}_i}, f_{{\text{R}}_i}^{\prime})$ denotes a pair of bearing vectors in the right camera. $ n_{\text{L}}$ and $n_{\text{R}}$ denote the number of features observed by the left and right camera. $\textbf{R}^{\text{b}}_{\text{c}_{\text{L}}}$ and $\textbf{R}^{\text{b}}_{\text{c}_{\text{R}}}$ represent the rotation matrix of the left camera-IMU extrinsic calibration and the rotation matrix of the right camera-IMU extrinsic calibration, respectively. $^\text{L}\textbf{M}_{\text{k},\text{k+1}} $ and $^\text{R}\textbf{M}_{\text{k},\text{k+1}}$ are formed by using the normals from the left and right cameras. $\mathcal{E}$ signifies a set of keyframe pairs at two consecutive times.

\subsection{Gyroscope Bias Refinement, Acceleration
Bias, Velocity and Gravity Estimator}
This step aims to attain optimal estimates of keyframes' velocities, gravity direction, and IMU biases using posterior estimation with the initial gyroscope bias from Step 1. The estimates from this step are as follows:
\begin{equation*}\small
\mathcal{X} = \left [\textbf{v}^{\text{w}}_{\text{b}_{0}:\text{b}_{N-1}}, \ \textbf{R}^{\text{w}}_{\text{g}}, \ \textbf{b}_{\text{g}}, \  \textbf{b}_{\text{a}}\right]^\top
\end{equation*}
and the posterior distribution is: %in this step can be formulated as follows:

\[p(\mathcal{X} | \mathcal{L}_{0:\text{k-1}}) \propto p(\mathcal{L}_{0:\text{k-1}}|\mathcal{X}) p(\mathcal{X})\]
where $\mathcal{L}_{0:\text{k}}$ refers to the preintegration of inertial measurements between consecutive keyframes, spanning from the first keyframe to the $\text{(k-1)}$-th keyframe, $p(\mathcal{L}_{0:\text{k-1}}|\mathcal{X})$ represents the likelihood of the inertial measurements given IMU states, and $p(\mathcal{X})$ is the prior of the IMU states.
The resulting optimal estimates $\mathcal{X}^*$ can be obtained by maximizing the posterior distribution, which is equivalent to minimizing its negative logarithm, leading to the following conversion:
\begin{equation}\small\label{eq:imuonly_map}
\begin{aligned}
 \mathcal{X}^* = \arg\max_{\mathcal{X}} p(\mathcal{X} | \mathcal{L}_{0:\text{k-1}}) 
 =\arg\min_{\mathcal{X}}\Big(-\log(p(\mathcal{X})) \\
-\sum_{k=0}^{N-2}\log\big(p(\mathcal{L}_{\text{k},\text{k+1}}|\textbf{R}^{\text{w}}_{\text{g}},  \textbf{b}_{\text{g}},  \textbf{b}_{\text{a}}, \textbf{v}^{\text{w}}_{\text{b}_{\text{k}}:\text{b}_{\text{k+1}}})\big)\Big) 
\\=
\arg\min_{\mathcal{X}}\Big({\Vert \textbf{r}_p\Vert}^2_{\Sigma_p} + \sum_{k=0}^{N-2}{\Vert\textbf{r}_{\mathcal{L}_{\text{k},\text{k+1}}}\Vert}^2_{\Sigma_{\mathcal{L}_{\text{k},\text{k+1}}}} \Big)  
\end{aligned}
\end{equation}
$\textbf{R}^{\text{w}}_{\text{g}}$ denotes the rotation aligning gravity with the world's z-axis( $\textbf{g}^{\text{w}} = \textbf{R}^{\text{w}}_{\text{g}}  \textbf{g}$).
$\textbf{r}_p$ and $\textbf{r}_{\mathcal{L}_{\text{k},\text{k+1}}}$ are the residuals of IMU biases prior and IMU measurements in two consecutive keyframes, and $\Sigma_p$ and $\Sigma_{\mathcal{L}_{\text{k},\text{k+1}}}$ are the corresponding covariances.
\begin{equation*}\small
\begin{aligned}
    \textbf{r}_{\mathcal{L}_{\text{k},\text{k+1}}} &= \left[ \delta{\boldsymbol\alpha}^{\text{b}_\text{k}}_{\text{b}_\text{k+1}}, \ \delta{\boldsymbol\beta}^{\text{b}_\text{k}}_{\text{b}_\text{k+1}}, \ \delta{\boldsymbol\gamma}^{\text{b}_\text{k}}_{\text{b}_\text{k+1}} \right]^\top \\
    &=\left[\begin{matrix} 
{\textbf{R}^{\text{w}}_{\text{b}_\text{k}}}^\top(\textbf{t}^{\text{w}}_{\text{b}_\text{k+1}} - \textbf{t}^{\text{w}}_{\text{b}_\text{k}} - \frac{1}{2}\textbf{g}^{\text{w}}\Delta \text{t}_{\text{k}, \text{k+1}}-\textbf{v}^{\text{w}}_{\text{b}_\text{k}}\Delta \text{t}_{\text{k}, \text{k+1}}) - \boldsymbol\alpha^{\text{b}_\text{k}}_{\text{b}_\text{k+1}} \\
{\textbf{R}^{\text{w}}_{\text{b}_\text{k}}}^\top(\textbf{v}^{\text{w}}_{\text{b}_\text{k+1}} -\textbf{g}^{\text{w}}\Delta \text{t}_{\text{k}, \text{k+1}} - \textbf{v}^{\text{w}}_{\text{b}_\text{k}}) - \boldsymbol\beta^{\text{b}_\text{k}}_{\text{b}_\text{k+1}} \\
\text{Log}(\boldsymbol\gamma^{\text{b}_\text{k}}_{\text{b}_\text{k+1}}
{\textbf{R}^{\text{w}}_{\text{b}_\text{k}}}^\top\textbf{R}^{\text{w}}_{\text{b}_\text{k+1}})
\end{matrix} \right]
\end{aligned}
\end{equation*}

\subsection{Rotation-Translation-Decoupled Optimization}
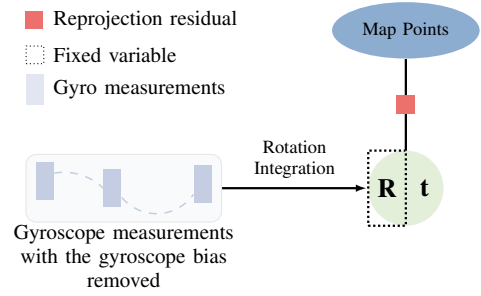
\begin{figure}[!ht]
    \vspace{0.21in}
	\centering
	\tikzsetnextfilename{translation_only}
	\definecolor{iceblue}{RGB}{140,172,208}
\definecolor{lightblue}{RGB}{196,205,225}
\definecolor{lightgreen}{RGB}{228,239,218}
\definecolor{lightred}{RGB}{236,112,107}

\begin{tikzpicture}
%	%\node [](Env) at (0.75, 2.0) {\includegraphics[width=.08\textwidth]{./figures/CARLA.png}};
%	\node [draw, rectangle, minimum width=1.2cm, minimum height = 0.8cm, font=\scriptsize, fill=yellow!40] at (3.1,2.0) (perception){Perception};
%	%\node[font=\scriptsize, align=center] at (1.2, 2.7) {Environment};
%	\node [draw, rectangle, fill=cyan!20, minimum width=1.2cm, minimum height = 0.8cm, font=\scriptsize, align=center] at (5.45,2.0)(RL){RL \\ Control};
%	\node [draw, rectangle, fill=blue!20, minimum width=1.2cm, minimum height = 0.8cm, font=\scriptsize, align=center] at (7.8,2.0)(Vehicle){Vehicle};
%	%
%	\draw [->, line width=0.3mm, >=latex]($(Env.east)+(-0.1,0)$) -- ($(perception.west) + (-0.0, 0) $) node[pos=0.45, above, font=\scriptsize]{Camera};
%	\draw [->, line width=0.3mm, >=latex](perception)--(RL) node[midway, above, font=\scriptsize]{Distance};
%	\draw [->, line width=0.3mm, >=latex](RL)--(Vehicle) node[midway, above, font=\scriptsize]{Brake};
%	\draw [->, line width=0.3mm, >=latex](Vehicle.south)-- ++(0.0, -0.5) -| (RL.south) node[pos=0.25, above, font=\scriptsize]{Velocity};
%	\draw [->,gray, line width=0.3mm, >=latex, dashed](Vehicle.east)-- ++(0.3, 0.0) -- ++(0.0, -1.2) -| (Env.south) node[pos=0.25, above, font=\scriptsize]{};
[
cell1/.style={draw=none, fill=lightred, rectangle, thick, minimum height=2mm, minimum width=2mm},
cell2/.style={draw, fill=none,  densely dotted, rectangle, minimum height=2mm, minimum width=2mm},
cell3/.style={draw=none, fill=lightblue, fill opacity=0.5, rectangle, minimum height=4.0mm, minimum width=2.0mm},
]
	\node[ellipse, draw=none, fill=iceblue, minimum width=1.5cm, minimum height=0.75cm, font=\scriptsize] (e) at (6, 2.5) {Map Points};
	\node[circle, draw=none, fill=lightgreen, minimum size=1.0cm] (c) at (6, 0.4) {};
		\node[draw, fill=lightblue, minimum height=0.9cm, minimum width=2.6cm, rounded corners, opacity=0.1] (r4) at (2.25, 0.4) {};
	\node[draw=none, fill=lightblue, minimum height=0.5cm, minimum width=0.25] (r1) at (1.2, 0.5) {};
	\node[draw=none, fill=lightblue, minimum height=0.5cm, minimum width=0.25] (r2) at (2.1, 0.4) {};
	\node[draw=none, fill=lightblue, minimum height=0.5cm, minimum width=0.25] (r3) at (3.3, 0.45) {};
	\node[draw, fill=none, line width=0.3mm, minimum height=1cm, minimum width=0.5cm, densely dotted] (r5) at (5.75, 0.4) {};
	%\draw[->, line width=0.3mm, >=latex, arrow head=3mm] (r4.east) -- (r5.west) node[midway, above,  align=center, font=\scriptsize] {Rotation \\ Integration};
	\draw[-{Latex[length=1.5mm, width=1mm]}, line width=0.3mm] (r4.east) -- (r5.west) node[midway, above,  align=center, font=\scriptsize] {Rotation \\ Integration};
	%\draw[decoration={markings,mark=at position 1 with {\arrow[ultra thick, Latex]{Latex}}},
	%={decorate}, line width=0.3mm] (r4.east) -- (r5.west) node[midway, above,  align=center, font=\scriptsize] {Rotation \\ Integration};
	\draw[line width=0.3mm] (c.north) -- (e.south);
	\node[draw=none, fill=lightred, minimum height=0.25cm, minimum width=0.25cm] (r3) at (6, 1.5125) {};
	
	\draw [line width=0.2mm, dashed, lightblue](1.325, 0.6) arc (100:45:0.75);
	\draw [line width=0.2mm, dashed, lightblue](2.225, 0.275) arc (-140:-35:0.6);
	
	\node [] at (5.75, 0.4) {\textbf{R}};
	
	\node [] at (6.25, 0.4) {\textbf{t}};
		\matrix [draw=none, below left] at (4.0, 3.0) {
		\node [cell1, label=right:\footnotesize Reprojection residual] {}; \\
		\node [cell2, label=right:\footnotesize Fixed variable]{};\\
		\node [cell3, label=right:\footnotesize Gyro measurements] {}; \\
	};
	\node [align=center, font=\footnotesize] at (2.25, -0.5) {Gyroscope measurements\\with the gyroscope bias \\ removed};
	
	%\draw[help lines](0,0) grid (8,3);
\end{tikzpicture}
	\caption{ 
 Step 3:  Rotation update via IMU integration, followed by translation optimization using 3-DoF bundle adjustment.}
	\label{fig:translate_opt}
\end{figure}

As shown in \fig{fig:translate_opt}, we separate the optimization of rotation and translation. This decision stems from our observation that rotation acquired through IMU rotation integration, after removing the gyroscope bias, is more accurate than the rotation derived from pure visual SLAM (refer to the RRE column labeled 'W/O VI-BA' in Table~\ref{tab:exp1}). Consequently, after attaining optimal gyroscope bias from Step 2, we update each rotation $\textbf{R}_{\text{b}_{\text{k}}}^{\text{w}}$ within the sliding window via IMU rotation integration and optimize each translation ${\textbf{t}^{\text{c}_{\text{k}}}_{\text{w}}}$ with 3-DoF BA:
\begin{equation}\small\label{eq:translate_opt}
\begin{aligned}
\textbf{R}_{\text{c}_{\text{k}}}^{\text{w}} &=\textbf{R}_{\text{b}_{\text{k}}}^{\text{w}}
\textbf{R}_{\text{c}_\text{L}}^{\text{b}} \\
{\textbf{t}^{\text{c}_{\text{k}}}_{\text{w}}}^* = \arg\min_{\textbf{t}^{\text{c}_{\text{k}}}_{\text{w}}}&\sum_{i\in\mathcal{M}}\rho\big(\Vert\textbf{x}^{i}- \pi_{(\cdot)}(\textbf{R}^{\text{c}_{\text{k}}}_{\text{w}}\textbf{X}^{i}+\textbf{t}^{\text{c}_{\text{k}}}_{\text{w}})\Vert^2_{\Sigma}\big) 
\end{aligned}
\end{equation}
$\textbf{X}^{i}$ denotes a 3D point  in the world frame, obtained through triangulation. and $\textbf{x}^{i}$ represents its corresponding 2D feature. $\rho$ is the robust Huber cost function, $\pi_{(\cdot)}$ are reprojection functions (monocular $\pi_m$ and rectified stereo $\pi_s$), and $\Sigma$ corresponds to the covariance related to the scale level of the keypoints in the pyramid~\cite{ORB2}. 
\subsection{Joint Visual-Inertial Bundle Adjustment}
After optimizing the rotation ${\textbf{R}^{\text{c}_{\text{k}}}_{\text{w}}}^*$ and translation ${\textbf{t}^{\text{c}_{\text{k}}}_{\text{w}}}^*$ in Step 3, we evaluate the success of the initialization process. Our approach involves computing the average residual of the normal epipolar constraint:
\begin{align*}
{\bar{e}} &= \frac{1}{N}\sum_{\text{k}=0}^{N-1}  \bar{e}_{\text{k},\text{k+1}} \\
\bar{e}_{\text{k},\text{k+1}} &= \frac{1}{n}\sum_{i=1}^n(\left|\textbf{n}_i^\top{\textbf{t}^{\text{c}_{\text{k}}}_{\text{w}}}^*\right|) \\
\textbf{n}_i &= f_i \times \textbf{R}^{\text{c}_\text{L}}_{\text{b}}\hat{\boldsymbol\gamma}^{\text{b}_\text{k}}_{\text{b}_\text{k+1}}\textbf{R}^{\text{b}}_{\text{c}_\text{L}}f_i^{\prime} 
\end{align*}
where $\bar{e}_{\text{k},\text{k+1}}$ represents the average residual of the normal epipolar constraint from time $\text{k}$ to time $\text{k+1}$. $(f_i, f_{i}^{\prime})$ is a pair of bearing vectors visible to both left and right cameras. $N$  represents the number of keyframes in the sliding window during initialization and $n$ denotes the number of pairs of covisible bearing vectors at two consecutive keyframes. If ${\bar{e}}$ falls below a certain threshold, we consider the initialization successful and proceed with the application of VI-BA. This choice is made because the previous steps offer not only precise initial estimates to serve as seeds for joint VI-BA, accelerating its convergence, but also expedite the runtime of VI-BA.

\section{Experimental Results} \label{sec:results}
\subsection{Experimental Setup and Implementation}\label{sec:implementation}
The EuRoC dataset~\cite{Burri25012016} provides precise rotation and translation data for 11 MAV sequences, spanning various flight conditions. It includes synchronized visual-inertial sensor units with global shutter cameras and a MEMS IMU for angular rate and acceleration data. Camera intrinsic and camera-IMU extrinsic parameters are also available. All experiments are conducted on an Intel i9-10920X desktop with 64 GB of RAM. 
\begin{figure*}[!ht]
    \vspace{0.02in}
	\centering
	\tikzsetnextfilename{wo_viba}
	\input{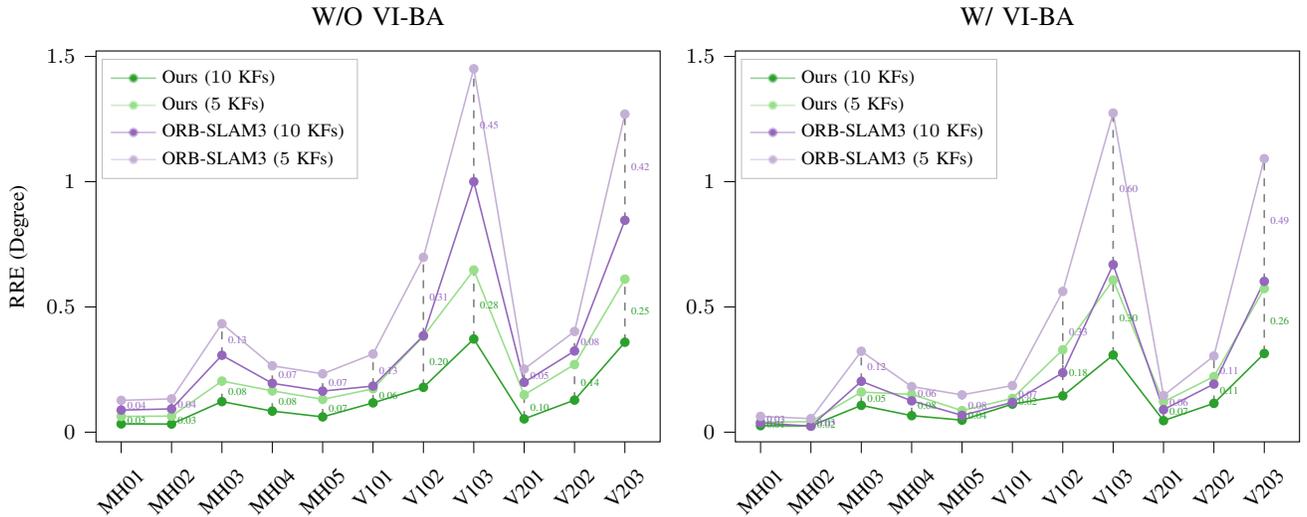}
	\caption{Relative Rotation Error (RRE) for different methods with different number of keyframes. \textbf{Left:} Results with 5 and 10 keyframes without  VI-BA for initialization.
		\textbf{Right:} Results with 5 and 10 keyframes with VI-BA applied for initialization.}
	\label{fig:exp3}
\end{figure*}        
To ensure a fair comparison between the initialization method of ORB-SLAM3~\cite{ORBSLAM3TRO} and Stereo-NEC, our method is integrated into ORB-SLAM3. 
In tackling \eq{eq:stereo_nec_bg}, quaternions serve as the chosen minimal rotation parameterization. $\textbf{M}_{\text{k},\text{k+1}}$ is a 3 $\times$ 3 matrix. \eq{eq:imuonly_map} and \eq{eq:translate_opt} can be solved iteratively using the Levenberg-Marquardt algorithm with analytic derivatives. ${\bar{e}} $ is set to $10^{-4}$ and the number of keyframes in the sliding window for initialization is 10.
In all our evaluation results, the absolute trajectory error (ATE) is measured in meters and does not involve scale alignment.
\begin{table*}[!ht]
\centering
{
\vspace{0.20in}
\caption{%\footnotesize 
Comparison of the accuracy of Stereo-NEC and ORB-SLAM3, both using 10 keyframes for initialization, with (W/) and without (W/O) VI-BA, in terms of ATE (meters) and RRE (degrees). Each value in the table corresponds to the average of RMSE results from different methods, which launch initialization every 2.5 seconds on the EuRoC dataset. 
 \\
}
\label{tab:exp1}
 \scalebox{0.88}{
\begin{tabular}{@{}lcccc|cccc}
\toprule  
\multirow{3}{*}{\textbf{Seq. Name}} &\multicolumn{4}{c|}{\textbf{ATE (m)}} & \multicolumn{4}{c}{\textbf{RRE (deg)}} \\
 &\multicolumn{2}{c}{\textbf{W/O VI-BA}} & \multicolumn{2}{c|}{\textbf{W/ VI-BA}} &\multicolumn{2}{c}{\textbf{W/O VI-BA}} & \multicolumn{2}{c}{\textbf{W/ VI-BA}}\\
&Ours& ORB-SLAM3 &Ours& ORB-SLAM3  &Ours& ORB-SLAM3  &Ours& ORB-SLAM3 \\
    \hline
     MH\_01\_easy &\textbf{0.005} &0.008&\textbf{0.005} &0.006&\textbf{0.033}&0.088 &\textbf{0.026}&0.037\\
     MH\_02\_easy &\textbf{0.005}&0.006&\textbf{0.004}&\textbf{0.004}&\textbf{0.032}&0.093&0.025&\textbf{0.024}\\
     MH\_03\_medium
    &\textbf{0.030}&0.032&\textbf{0.025}&0.027&\textbf{0.122}&0.307&\textbf{0.107}&0.203\\
     MH\_04\_difficult &\textbf{0.029}&0.031&\textbf{0.018}&0.026&\textbf{0.084}&0.195&\textbf{0.066}&0.125  \\
     MH\_05\_difficult &\textbf{0.020}&0.027&\textbf{0.014}&0.016&\textbf{0.061}&0.164&\textbf{0.048}&0.067\\
     V1\_01\_easy &\textbf{0.007}&0.008&\textbf{0.006}&\textbf{0.006}&\textbf{0.117}&0.184&\textbf{0.112}&0.119\\
     V1\_02\_medium &\textbf{0.018}&0.020&\textbf{0.012}&0.015&\textbf{0.179}&0.385&\textbf{0.145}&0.237\\
     V1\_03\_difficult &\textbf{0.040}&0.054&\textbf{0.026}&0.043&\textbf{0.372}&1.000&\textbf{0.308}&0.669\\
     V2\_01\_easy  &\textbf{0.004}&0.007&\textbf{0.003}&0.004&\textbf{0.053}& 0.199&\textbf{0.046}&0.090\\
     V2\_02\_medium &\textbf{0.013}&0.021&\textbf{0.009}&0.014&\textbf{0.128}&0.324&\textbf{0.115}&0.192\\
     V2\_03\_difficult &\textbf{0.036}&0.047&\textbf{0.028}&0.039&\textbf{0.359}&0.846&\textbf{0.314}&0.602\\
    \hline
    Avg &\textbf{0.019} &0.024&\textbf{0.014} &0.018&\textbf{0.140}&0.344&\textbf{0.119}&0.215\\
\bottomrule
\end{tabular}}
}
\end{table*}

\subsection{Accuracy Evaluation} \label{sec:accuracy_performance}
To measure accuracy on different trajectories, we perform an exhaustive initialization test. During this test, we launch an initialization with 10 keyframes every 2.5 seconds for each sequence, leading to the evaluation of 464 different initialization segments. 
% The choice of a 2.5-second interval is based on the fact that all the initialization methods we compared required approximately 2 seconds to run trajectories and perform pure visual SLAM with 10 keyframes, which takes around 500 milliseconds. 
Table~\ref{tab:exp1} presents a comparison of the absolute trajectory error (ATE) and relative rotation error (RRE) for ORB-SLAM3 and Stereo-NEC, with and without visual-inertial bundle adjustment (VI-BA). 
Across various sequences, our method consistently outperforms ORB-SLAM3 in terms of both ATE and RRE with the exception of RRE in MH\_02\_easy, for which the two methods are very closely matched.
On average, our method achieves 1.3 times better accuracy in ATE without utilizing  VI-BA, compared to ORB-SLAM3. Furthermore, there is an impressive reduction of RRE  by a factor of 2.5. When VI-BA is employed, the performance of our method is further improved: ATE drops to 0.014 meters, and RRE reduces to 0.119 degrees. In contrast, under the same conditions, ORB-SLAM3 registers an ATE of 0.018 meters and an RRE of 0.215 degrees.

The extensive results demonstrate that our approach, which involves integrating bias-removed gyroscope measurements for camera rotation updates, stands in contrast to ORB-SLAM3's reliance solely on camera rotation estimation from pure visual SLAM. This contrast leads to substantial improvements in rotation estimation. Moreover, the combination of precise rotation estimates and our translation-only optimization further enhances translation estimation.
% \begin{figure*}[!ht]
%         \vspace{0.02in}
% 	\centering
% 	\tikzsetnextfilename{wo_viba}
% 	\input{figures/wo_viba}
% 	\caption{Relative Rotation Error (RRE) for different methods with different number of keyframes. \textbf{First column:} Results with 5 and 10 keyframes without  VI-BA for initialization.
% 		\textbf{Second column:} Results with 5 and 10 keyframes with VI-BA applied for initialization.}
% 	\label{fig:exp3}
% \end{figure*}

\begin{table*}[!ht] 
\centering
{
\vspace{0.10in}
\caption{%\footnotesize 
Comparison of average initialization computation time for 10 keyframes setting in milliseconds (ms) on EuRoC.\\
The results involve 10 keyframes, and the best results for each sequence are highlighted in bold.
}
\label{tab:exp5}
 \scalebox{0.79}{
\begin{tabular}{@{}lc|cccc|cccc}
\toprule  
\multirow{2}{*}{\textbf{Seq. Name}}&{\textbf{Ours} \& \textbf{ORB-SLAM3}} &\multicolumn{4}{c|}{\textbf{Ours}} & \multicolumn{4}{c}{\textbf{ORB-SLAM3}} \\
&Pure Visual SLAM& Bias, Vel \& Grav Est&Rot-Int \& Trans-Opt& VI-BA  &Total Cost& Bias, Vel \& Grav Est&Rot-Int \& Trans-Opt& VI-BA&Total Cost \\
    \hline
     MH\_01\_easy &514.84 &291.58 &15.55 &\textbf{57.06} &879.03&\textbf{1.97}&-&68.23&\textbf{585.04}\\
     MH\_02\_easy &474.32 &311.86 &17.53 &\textbf{58.31} &862.02&\textbf{1.95}&-&68.56&\textbf{544.83}\\
     MH\_03\_medium &458.50 &292.21 &16.37 &\textbf{56.74} &823.82&\textbf{2.10}&-&62.83&\textbf{523.43}\\
     MH\_04\_difficult &466.76 &292.73 &14.57 &\textbf{52.33} &826.39&\textbf{2.22}&-&66.56&\textbf{535.54}\\
     MH\_05\_difficult &472.50 &308.98 &17.15 &\textbf{52.49} &851.12&\textbf{2.06}&-&64.21&\textbf{538.77}\\
     V1\_01\_easy &563.99 &304.82 &22.79 &\textbf{74.32} &965.92&\textbf{1.80}&-&77.42&\textbf{643.21}\\
     V1\_02\_medium &495.96 &298.41 &16.43 &\textbf{53.21} &864.01&\textbf{1.79}&-&60.86&\textbf{558.61}\\
     V1\_03\_difficult & 532.10&252.32 &14.69 &\textbf{49.81} &848.92&\textbf{2.05}&-&60.30&\textbf{594.45}\\
     V2\_01\_easy  &553.45 &331.41 &20.75 &\textbf{73.60} &979.21&\textbf{1.90}&-&80.78&\textbf{636.13}\\
     V2\_02\_medium &556.98 &300.84 &14.52 &\textbf{52.20} &924.54&\textbf{2.06}&-&62.10&\textbf{621.14}\\
     V2\_03\_difficult & 484.68&264.52 &12.51 &\textbf{40.54} &802.25&\textbf{2.11}&-&57.70&\textbf{544.49}\\
    \hline
    Avg &506.73 &295.43 &16.62 &\textbf{56.42}  &875.20&\textbf{2.0}&-&66.32&\textbf{575.05}\\
\bottomrule
\end{tabular}}
}
\end{table*}

\subsection{Robustness Evaluation} \label{sec: robust performance}
%We conduct two separate evaluations to assess the robustness metrics of different methods. The results are presented in two sections, showcasing the outcomes with and without VI-BA. These evaluations provide valuable insights into the methods' robustness under various conditions and initialization settings.
We conduct two separate evaluations, each with and without VI-BA, to assess the methods' robustness under various conditions and initialization settings.

%\textbf{First Evaluation}:
\noindent\textbf{Reduced number of keyframes:} We investigate how a more robust method would demonstrate a small decrease in RRE when initialized with fewer keyframes. The results are depicted in~\fig{fig:exp3} which shows the RRE when using 5 or 10 keyframes for initialization.

When transitioning from 10 to 5 keyframes for initialization without utilizing VI-BA, our method experiences an average increase of 0.119 degrees in RRE. ORB-SLAM3 shows a slightly higher increase of 0.163 degrees under the same conditions. On the other hand, when our method employs VI-BA, the RRE increases by a modest average of 0.105 degrees, while ORB-SLAM3 exhibits a larger increase of 0.179 degrees. Notably, in scenes with large rotation (V1\_03 difficult and V2\_03 difficult), regardless of VI-BA usage, ORB-SLAM3 demonstrates RRE changes 1.7-2.0 times higher than those of our method. Moreover, across all sequences, our method achieves lower RRE, despite utilizing fewer keyframes (5 keyframes) for initialization, compared to ORB-SLAM3's RRE when using 10 keyframes. This observation serves as strong evidence for the precision and robustness of our approach. 

%\textbf{Second Evaluation}: 
\noindent\textbf{Challenging conditions:} We test the methods under challenging conditions, such as motion blur and illumination changes, using segments from V2\_03\_difficult. To do this, we categorize the data segments into three groups based on the 10 keyframes' average angular velocity magnitude $\bar{\omega}$: low-speed ($5^{\circ}$/s $\leq$ $\parallel\bar{\omega}\parallel$ $<$ $15^{\circ}$/s), medium-speed ($15^{\circ}$/s $\leq$ $\parallel\bar{\omega}\parallel$ $<$ $30^{\circ}$/s), and high-speed  ($\parallel\bar{\omega}\parallel$ $\geq$ $30^{\circ}$/s) data segments. The results are presented in Table \ref{tab:exp2}.

On average, our method achieves an ATE of 0.035 meters and an RRE of 0.312 degrees without VI-BA, whereas ORB-SLAM3 achieves an ATE of 0.044 meters and an RRE of 0.706 degrees. When utilizing VI-BA, the ATE of our method decreases to 0.028 meters and the RRE to 0.268 degrees, while the ATE of ORB-SLAM3 remains at 0.035 meters and the RRE at 0.446 degrees. Specifically, in high-speed scenarios, ORB-SLAM3 demonstrates 2.4-3.1 times higher RRE and 2.0-2.2 times higher ATE, regardless of whether VI-BA is used or not. These findings highlight the improved performance and robustness of our method compared to ORB-SLAM3 in high-speed scenarios.
\begin{table}[!tbp] %[!tbp]
\centering
{
\vspace{0.20in}
\caption{%\footnotesize 
Exhaustive initialization results for 10 keyframes with Low, Medium, and High Angular Velocity from V2\_03\_difficult sequence.
}
\label{tab:exp2}
 \resizebox{\columnwidth}{!}{
\begin{tabular}{@{}lcccc|cccc}
\toprule  
\multirow{3}{*}{\textbf{Seq. Name}} &\multicolumn{4}{c|}{\textbf{ATE (m)}} & \multicolumn{4}{c}{\textbf{RRE (degree)}} \\
 &\multicolumn{2}{c}{\textbf{W/O VI-BA}} & \multicolumn{2}{c|}{\textbf{W/ VI-BA}} &\multicolumn{2}{c}{\textbf{W/O VI-BA}} & \multicolumn{2}{c}{\textbf{W/ VI-BA}}\\
&Ours& ORB-SLAM3 &Ours& ORB-SLAM3  &Ours& ORB-SLAM3  &Ours& ORB-SLAM3 \\
    \hline
     Low &\textbf{0.023}&0.028&0.017&\textbf{0.015}&\textbf{0.151}&0.378&0.133&\textbf{0.085}\\
     Medium &\textbf{0.059}&0.059&\textbf{0.050}&0.051&\textbf{0.465}&0.737&\textbf{0.362}&0.513\\
    High &\textbf{0.023}&0.045&\textbf{0.017}&0.038&\textbf{0.319}&1.003&\textbf{0.308}&0.741\\
    \hline
    Avg &\textbf{0.035} &0.044&\textbf{0.028} &0.035&\textbf{0.312}&0.706&\textbf{0.268}&0.446\\
\bottomrule
\end{tabular}}
}
\end{table}

\subsection{Computation Speed Evaluation} \label{sec: speed performance}
In Table~\ref{tab:exp5}, we present the runtime comparison for each  initialization module separately, including pure visual SLAM, IMU bias, velocity and gravity estimation (Bias, Vel \& Grav Est), rotation integration, and translation optimization (Rot-Int \& Trans-Opt) and VI-BA.

The results reveal that our method is around 10 ms faster on average than ORB-SLAM3 in VI-BA because it provides accurate initial estimates which aid faster convergence. However, our method takes 300.15 milliseconds longer on average for initialization compared to ORB-SLAM3. This is due to two additional steps in our method: 1) We first estimate the initial gyroscope bias before estimating keyframes' velocities, gravity direction, and acceleration bias, while ORB-SLAM3's Inertial-only step simultaneously estimates velocities, gravity direction, and IMU biases. 2) After obtaining the gyroscope bias, we refine the camera rotation estimation by integrating gyroscope measurements with the gyroscope bias removed, and we update the camera translation using a 3-DoF bundle adjustment, while ORB-SLAM3 does not update the camera pose from pure visual SLAM.

These two additional steps are indispensable, especially when dealing with a larger gyroscope bias IMU. Enhancing pose estimation has been demonstrated by updating the camera rotation through the integration of gyroscope bias-removed measurements and the camera translation via 3-DoF bundle adjustment with updated rotation estimation, as shown in ~\ref{sec:accuracy_performance}. 
%Additionally, both methods require running a trajectory for about 2.3 seconds, making 300 milliseconds insignificant in comparison to the 2.3 seconds duration.
The additional 300 ms are only required during initialization and can be considered negligible for trajectories that last even a few seconds.          
\section{CONCLUSIONS}
Our proposed method, Stereo-NEC, addresses limitations in the current state-of-the-art approach, ORB-SLAM3, which heavily relies on the accuracy of pure visual SLAM to estimate inertial variables without initial gyroscope bias estimation in inertial-only optimization. We independently estimate the gyroscope bias, then use it to refine other parameters through a maximum a posteriori problem. After this, we update rotation estimation via IMU integration with the gyroscope bias removed, enhancing translation estimation through 3-DoF bundle adjustment with updated rotation estimation. We also introduce a novel approach to determine initialization success by evaluating the residual of the normal epipolar constraint. As a result, our method improves both accuracy and robustness compared to ORB-SLAM3, %demonstrating superior performance in terms of absolute trajectory error and relative rotation error, 
while maintaining competitive computation speed.
\balance
\printbibliography

\end{document}